\def\year{2020}\relax
\documentclass[letterpaper]{article} 
\usepackage{aaai20}  
\usepackage{times}  
\usepackage{helvet} 
\usepackage{courier}  
\usepackage[hyphens]{url}  
\usepackage{graphicx} 
\usepackage{amsmath}
\usepackage{amsfonts}
\usepackage{diagbox}
\usepackage{multirow}
\usepackage{booktabs}
\usepackage{algorithm}
\usepackage{algorithmic}
\usepackage{enumerate}
\usepackage{bm}
\usepackage{subcaption}
\nocopyright
\newcommand{\mb}{\mathbb}
\newcommand{\mr}{\mathrm}
\newcommand{\mc}{\mathcal}

\newcounter{rek_counter}

\newtheorem{remark}[rek_counter]{Remark}

\title{Visual Tactile Fusion Object Clustering}

\author{Tao Zhang,\textsuperscript{\rm 1,2} Yang Cong,\textsuperscript{\rm 1}\thanks{The corresponding author is Prof. Yang Cong.} Gan Sun,\textsuperscript{\rm 1,2}\thanks{The author contributed equally to this work.} Qianqian Wang,\textsuperscript{\rm 3} Zhenming Ding\textsuperscript{\rm 4}\\
\textsuperscript{\rm 1}State Key Laboratory of Robotics, Shenyang Institute of Automation, Chinese Academy of Sciences.\thanks{This work is supported by NSFC (61821005, 61722311, U1613214, 61533015), and LiaoNing Revitalization Talents Program(XLYC1807053).}\\
\textsuperscript{\rm 2}University of Chinese Academy of Sciences, \textsuperscript{\rm 3}Xidian University,
\textsuperscript{\rm 4}Indiana University-Purdue University Indianapolis, USA\\
zhangtaosia@gmail.com, congyang81@gmail.com, sungan1412@gmail.com, qianqian174@foxmail.com, zd2@iu.edu
}
\begin{document}

\maketitle

\begin{abstract}
Object clustering, aiming at grouping similar objects into one cluster with an unsupervised strategy, has been extensively-studied among various data-driven applications.
However, most existing state-of-the-art object clustering methods (e.g., single-view or multi-view clustering methods) only explore visual information, while ignoring one of most important sensing modalities, i.e., tactile information which can help capture different object properties and further boost the performance of object clustering task.
To effectively benefit both visual and tactile modalities for object clustering, in this paper, we propose a deep Auto-Encoder-like Non-negative Matrix Factorization framework for visual-tactile fusion clustering.
Specifically, deep matrix factorization constrained by an under-complete Auto-Encoder-like architecture is employed to jointly learn hierarchical expression of visual-tactile fusion data, and preserve the local structure of data generating distribution of visual and tactile modalities.
Meanwhile, a graph regularizer is introduced to capture the intrinsic relations of data samples within each modality.
Furthermore, we propose a modality-level consensus regularizer to effectively align the visual and tactile data in a common subspace in which the gap between visual and tactile data is mitigated.
For the model optimization, we present an efficient alternating minimization strategy to solve our proposed model.
Finally, we conduct extensive experiments on public datasets to verify the effectiveness of our framework.
\end{abstract}

\section{Introduction}
\begin{figure}[!ht]
\centering
\centerline{\includegraphics[width =.95\columnwidth]{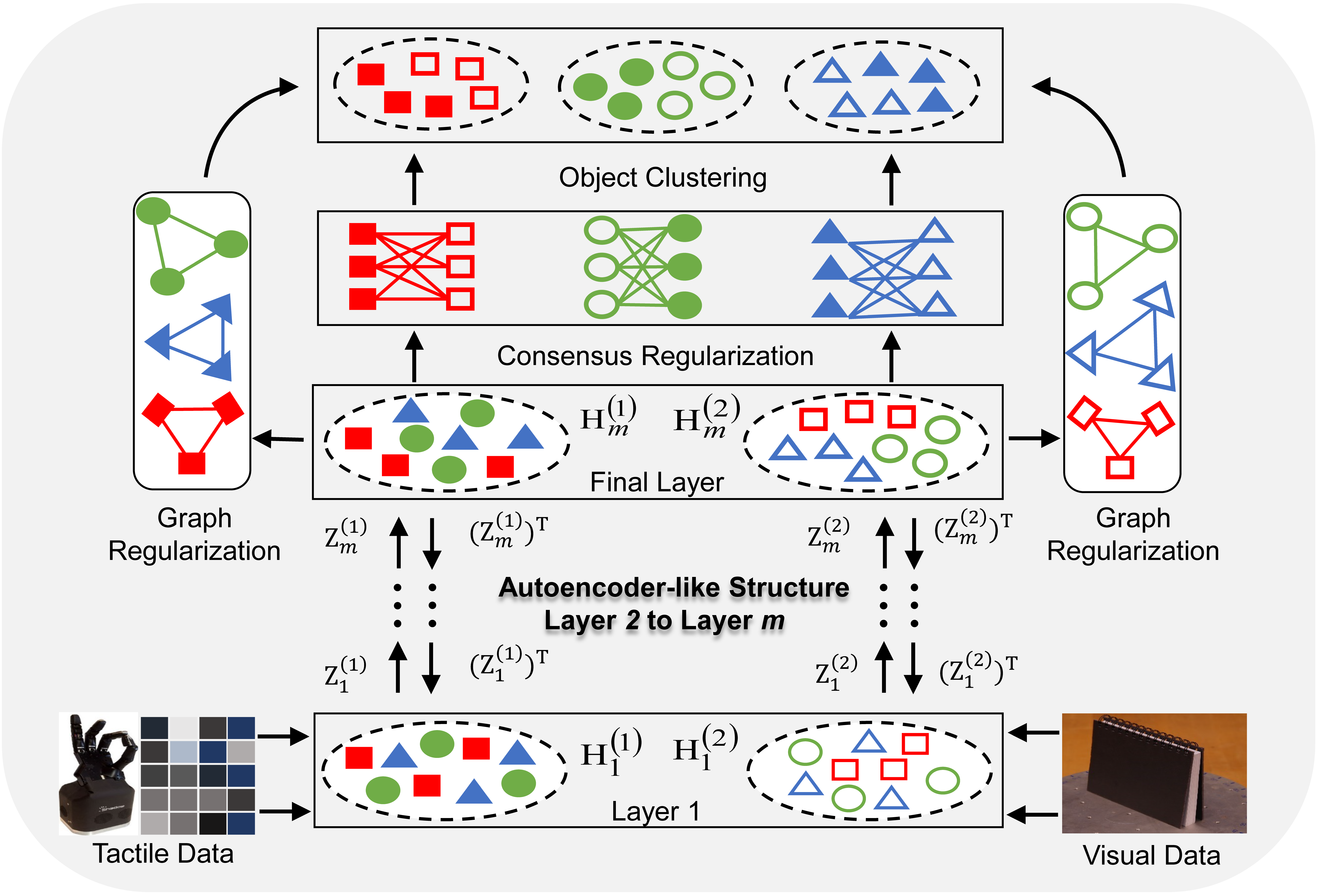}}
\caption{Illustration of the proposed visual-tactile fusion object clustering framework, where an under-complete Auto-Encoder-like structure is used to preserve the local structure of data generating distribution of visual and tactile modalities. With the consensus regularization, the gap between visual and tactile modalities can be well mitigated.
}
\label{fig:1}
\end{figure}

Grouping a set of objects in an unsupervised way that objects in the same group (called a cluster) are more similar to each other than these in other groups (i.e., object clustering) has attracted a lot attention in both academic and industrial communities in the past decades.
Most current object clustering works ~\cite{cao2015diversity,zhao2017multi,zhang2018generalized,zhang2018binary,yang2019deep,gan2020lifelong} aim at recognizing ``similar behavior'' based on visual information captured by a visual camera (e.g., RGB or Depth camera) or represented by different description methods (e.g., SURF, LBP or deep features).
These above methods have been successfully applied into statistics, computer vision, biology or psychology~\cite{wu2013constrained,li2017sparse,li2017projective,liu2018multi,sun2019representative}.

However, most existing object clustering works ignore one of the important sensing modality, i.e., tactile information (e.g., hardness, force, and temperature), which casts a light in compensating visual information on many practical manipulation tasks~\cite{liu2016visual,yuan2017connecting}.
For example, in the practical situation that a robot grasps an apple, the visual information of the apple becomes unobservable due to the occlusion of a robot hand while the tactile information can be easily obtained. Some objects whose appearance are visually similar can be hardly distinguished via merely using visual information (e.g., ripe versus unripe fruits). However, the ripe versus unripe fruits can be easily distinguished by tactile properties (e.g., hardness). Besides, some objects cannot be well distinguished only by either visual information or tactile information. For instance, it is hard to differentiate three visually similar bottles, where two bottles are empty and the remaining one is full of water. Hence, it is beneficial from each other to perform object clustering by fusing visual and tactile modalities.

To integrate visual with tactile information, a naive solution is to treat visual or tactile data as single view data, and directly perform the existing multi-view clustering methods on the visual-tactile object clustering task.
However, the gap between visual and tactile modalities is very large~\cite{liu2018robotic}.
On the one hand, the devices which are used to collect tactile and visual data are different.
Tactile sensor obtains tactile data through constant physical contact, while the visual modality
can simultaneously generate multiple different features of an object at a distance.
Moreover, the format, frequency and receptive field is diverse since visual sensor usually perceives color, global shape and rough texture, while touch sensor is usually used to acquire detailed texture, hardness and temperature.
Therefore, how to establish a novel visual-tactile fusion object clustering model, which can tackle intrinsic gap challenge across visual and tactile data, is our focus in this work.

To address the challenges mentioned above, in this paper, we propose a deep Auto-Encoder-like Non-negative Matrix Factorization (NMF) framework for visual-tactile fused object clustering.
More specifically, deep NMF constrained with an under-complete Auto-Encoder-like structure is adopted to learn the hierarchical semantics, while preserving the local data structure among visual and tactile data in a layer-wise manner.
Then, we introduce a graph regularizer to reduce the differences between similar points inside each modality.
Furthermore, as a non-trivial contribution, we carefully design a sparse consensus regularizer to tackle the intrinsic gap problems between visual and tactile data.
We explore a consensus constraint to interact the individual component between different modalities with final consensus representation to align two modalities. Thus, it plays as the modality-level constraint to supervise the generation of a common subspace, in which the mutual information on visual and tactile data is maximized. To optimize our proposed framework, an efficient alternating minimization strategy is present. To the end, we conduct extensive experiments on public datasets to evaluate the effectiveness of our framework, wherein ours outperforms the state-of-the-arts. The contributions are summarized as:
\begin{itemize}
\item We propose a deep Auto-Encoder-like Nonnegative Matrix Factorization framework for visual-tactile fusion object clustering.
    To our best knowledge, this is a pioneering work to incorporate visual modality with tactile modality in the object clustering task.
\item We develop an under-complete Auto-Encoder-like structure to jointly learn the hierarchical semantics and preserve the local data structure.
    Meanwhile, we design a sparse consensus regularization to seek a common subspace, in which the gap between visual and tactile modalities is mitigated and the mutual information is maximized.
\item To solve our proposed framework, an efficient solution based on an alternating direction minimization method is provided. Extensive experiment results verify the effectiveness of our proposed framework.
\end{itemize}

\section{Related Work}
The work in this paper lies in the tasks of visual-tactile sensing and multi-view clustering. We thus introduce the related work including visual-tactile sensing and multi-view clustering in this section.

\subsection{Visual-Tactile Sensing}
Vision and touch are the most important sensing modalities both for robots and humans, and they are widely-applied in robot tasks~\cite{ilonen2014three,liu2016visual,yuan2017connecting,Dong_2019_ICCV}. Generally, visual-tactile sensing can be mainly divided into three categories including object recognition, 3D reconstruction and cross-modal matching.

Amongst the fields mentioned above, Liu et al. propose a visual-tactile fusion framework to recognize household objects based on kernel sparse coding method~\cite{liu2016visual}. Yuan and Luo et al. propose a deep learning framework for clothing material perception by fusing visual and tactile information~\cite{yuan2018active}.
Ilonen et al. develop to reconstruct 3D model of unknown symmetric objects by fusing visual and tactile information~\cite{ilonen2014three}. Wang et al. present to perceive accurate 3D object shape with a monocular camera and a high-resolution tactile sensor~\cite{wang20183d}.
Yuan et al. propose a multi-input net to connect the visual and tactile properties of fabrics~\cite{yuan2017connecting}.
Li et al. introduce a conditional generative adversarial network based prediction model to connect visual and tactile measurement~\cite{li2019connecting}.
Although the previous models have been successfully applied in supervised learning in the visual-tactile sensing fields, its application in object clustering is still under insufficient exploration.

\subsection{Multi-View Clustering}
Multi-view clustering has shown remarkable successes in many real-world applications. Based on standard spectral clustering~\cite{ng2002spectral}, co-training~\cite{kumar2011coT} and co-regularizer~\cite{kumar2011coR} are performed to enforce consistence of different views.
Based on the subspace clustering strategy, Cao and Zhang et al. try to capture complementary information from different views in the manner of subspace representations~\cite{cao2015diversity,zhang2018generalized} .
Based on the framework of non-negative matrix factorization and its variants~\cite{trigeorgis2014deep},
Li et al. propose a consensus clustering and semi-supervised clustering method based on Semi-NMF~\cite{li2007solving}.
Zhao et al. propose a deep Semi-NMF method for multi-view clustering~\cite{zhao2017multi}.

\section{The Proposed Method}
\subsection{NMF Revisit}
NMF and its variants~\cite{lee2001algorithms,liu2011constrained} have previously shown to be promising in the field of multi-view clustering. The objective of NMF can be defined as:
\begin{equation}\label{eq:semiNMF}
\min \limits_{Z\geq 0,H \geq 0} \| X - ZH \|^2_{F},
\end{equation}
where $X$ is the input feature matrix, $Z$ is the basis matrix and $H$ is the compact representation, respectively.
We can obtain the final clustering result by performing standard spectral clustering~\cite{ng2002spectral} on $H$.
However, in real-world applications, it is not enough to learn intrinsic data structure with single-layer NMF due to complex data structure and data noise.
Zhao et al. show that a deep NMF model has an appealing performance in data representation~\cite{zhao2017multi}.
The deep NMF can be formulated as:
\begin{equation}\label{eq:DsemiNMF}
\begin{aligned}
X \approx Z_1H_1,  ~~\cdots,~~ X \approx Z_1\ldots Z_mH_m,
\end{aligned}
\end{equation}
where $Z_i$ and $H_i$ represent the basis matrix and representation for the $i$-th layer, respectively.
Inspired by this idea, we intend to explore the deep NMF architecture into our visual-tactile object clustering framework.

\subsection{The Proposed Framework}
In the setting of visual-tactile fusion object clustering framework, we use $X = \{ X^{(1)},\ldots,X^{(v)},\ldots,X^{(V)}\}$ as the input data, where $V$ is the number of modalities ($V$ is defined as $2$ for the visual-tactile clustering task in this work), and $v$ represents the $v$-th modality. $X^{(v)} \in \mb{R}^{d_v\times n}$ denotes the feature matrix for the $v$-th modality, $d_v$ represents the dimension of the feature, $n$ denotes the number of data samples. Then, we propose our deep visual-tactile fused object clustering model as follows:
\begin{footnotesize}
\begin{equation}
\label{eq:frameworkDefine}
\begin{aligned}
\min \limits_{Z_i^{(v)},H_i^{(v)} \atop H_m^{(v)}, H^{\ast}} &  \sum_{v=1}^V\big( \| X^{(v)}-Z^{(v)}_{1}Z^{(v)}_{2}\ldots Z^{(v)}_{m}H^{(v)}_{m} \|^2_{F} \\
& \! +\!  \| H^{(v)}_{m}-(Z^{(v)}_{m})^{\top}(Z^{(v)}_{m-1})^{\top}\ldots (Z^{(v)}_{1})^{\top}X^{(v)} \|^2_{F}  \\
& \!+ \! \beta  \mr{tr}(H^{(v)}_{m}L^{(v)} (H^{(v)}_{m})^{\top} ) \big) \!+\! \lambda \! \sum_{v=1}^V \! \| H^{(v)}_{m}\! -\! G^{(v)}H^{\ast}\|_{2,1}\\
s.t., &~H^{(v)}_{i} \geq 0, Z^{(v)}_{i} \geq 0, i = 1,2\ldots m,
\end{aligned}
\end{equation}
\end{footnotesize}
where $m$ is the number of layer, $\beta \geq0$ and $\lambda \geq0$ are the regularization parameters. $H^{(v)}_{m}$ represents the high hierarchical semantics of the $v$-th modality.

Moreover, the first and second terms denote the NMF constrained by an under-complete Auto-Encoder-like structure, which is designed to learn the hierarchical semantics while preserving the local structure of the input visual and tactile data.
The first term denotes an under-complete decoder process controlling the dimension of $H^{(v)}_{m}$ lower than $ X^{(v)}$ and further force NMF to learn more salient features representation of $ X^{(v)}$. The second term denotes an encoder process which implicitly maintains the local data structure via recovering $ X^{(v)}$ from $H^{(v)}_{m}$.
Furthermore, we have the following \textbf{Remarks} for the used regularization.
\begin{remark}
The graph regularization in the third term is designed to pull the similarities of nearby points inside each modality. $L^{(v)}$ denotes the graph Laplacian matrix for the $v$-th modality, constructed in $k$-nearest neighbor manner. By using the Eigen-decomposition technique on $L^{(v)}$, i.e., $L^{(v)} = Q^{(v)}P^{(v)}{(Q^{(v)})}^{\top}$, we obtain: $\mr{tr}(H^{(v)}_{m}L^{(v)} (H^{(v)}_{m})^{\top}) = \left\| H_{m}^{(v)}A^{(v)} \! \right\|_{F}^2$, where $A^{(v)} = Q^{(v)}{P^{(v)}}^{\frac{1}{2}}$. However, the process of collecting tactile or visual data is easily contaminated by environmental change, which leads to noise and outliers in the source data. Meanwhile, Frobenius norm is sensitive to the noises and outliers. We thus replace Frobenius norm by the $\ell_{2,1}$-norm, which can jointly remove outliers and uncover more shared representation across the nearby points inside each modality.
\end{remark}

\begin{remark}
  The last item is the consensus regularization, which is designed to tackle the intrinsic gap problem between visual and tactile data. This term directly measures the similarity between $H^\ast$ and $H_{m}^{(v)}$ in a utility way, where $G^{(v)}$ is the best mapping matrix to align $H^{(v)}_m$ to $H^\ast$.   After  aligning $H^{(v)}_m$ to $H^\ast$, the $\ell_{2,1}$-norm constraint is to calculate the dissimilarity between $H^{(v)}_m$ and $H^\ast$ in an efficient way.
  Therefore, this term plays as a modality-level constraint and learn a project matrix $G^{(v)}$, which projects $H^{(v)}_m$ into the common subspace $H^\ast$. In this subspace, the mutual information on each modality is maximized, which ultimately contributes to the object clustering.
\end{remark}

Then the objective function Eq.~\eqref{eq:frameworkDefine} is further reformulated as:
\begin{footnotesize}
\begin{equation}\label{eq:frameworkDetail}
\begin{aligned}
\min \limits_{Z_i^{(v)},H_i^{(v)} \atop H_m^{(v)}, H^{\ast}} &  \sum_{v=1}^V\big( \| X^{(v)}-Z^{(v)}_{1}Z^{(v)}_{2}\ldots Z^{(v)}_{m}H^{(v)}_{m} \|^2_{F}  \\
&  +  \| H^{(v)}_{m}-(Z^{(v)}_{m})^{\top}(Z^{(v)}_{m-1})^{\top}\ldots (Z^{(v)}_{1})^{\top}X^{(v)} \|^2_{F}  \\
& +  \beta  \|H^{(v)}_{m}A^{(v)}\|_{2,1} + \lambda  \| H^{(v)}_{m} - G^{(v)}H^{\ast}  \|_{2,1} \big)          \\
s.t., &~H^{(v)}_{i} \geq 0, Z^{(v)}_{i} \geq 0, i = 1,2\ldots m.
\end{aligned}
\end{equation}
\end{footnotesize}

\subsection{Optimization}
To efficiently solve the optimization problem Eq.~\eqref{eq:frameworkDetail}, we propose a solution based on alternating direction minimization algorithm.
To reduce the training time, we pre-train each layer to approximate the factor matrices $Z^{(v)}_i$ and $H_i^{(v)}$.
For the pre-training process, we decompose the input data matrix $X^{(v)}\approx Z^{(v)}_1H^{(v)}_1$ by minimizing $ \| X^{(v)}-Z^{(v)}_{1}H^{(v)}_{1} \|^2_{F} +  \| H^{(v)}_{1}-(Z^{(v)}_{1})^{\top}X^{(v)}\|^2_{F}$ first, where $Z^{(v)}_1 \in \mb{R}^{d_v\times p_1} $ and $H^{(v)}_1 \in \mb{R}^{p_1\times n} $. Then we decompose $H^{(v)}_1$ as $H_{1}^{v}\approx Z_{2}^{(v)}H_{2}^{(v)}$, where $Z^{(v)}_2 \in \mb{R}^{d_v\times p_2} $ and $H^{(v)}_2 \in \mb{R}^{p_2\times n}$.
$p_1$ is the dimension of layer $1$ and $p_2$ is the dimension of layer $2$\footnote{The layer size for layer $1$ to $m$ is denoted as $[p_1\ldots p_m]$ in this paper}. Repeating the process until all layers have been pre-trained. Then each layer is fine-tuned by alternating minimization of the proposed framework in Eq.~\eqref{eq:frameworkDetail}. Specifically, the update rules for each variable are as follows.

\subsubsection{Update rule for {$Z^{(v)}_i$}:}
With other variables fixed, we can have the following Lagrangian objective function:
\begin{footnotesize}
\begin{equation}\label{eq:Z}
\begin{aligned}
&  \min \limits_{Z_{i}^{(v)}}  \sum_{v=1}^V(\| X^{(v)}\!-\!\Phi Z^{(v)}_{i}H^{(v)}_{i} \|^2_{F} \!+\! \| H^{(v)}_{i}\!-\!(Z^{(v)}_{i})^{\top}\Phi^{\top}X^{(v)} \|^2_{F}\\
& \quad\quad- \eta Z^{(v)}_{i}),
\end{aligned}
\end{equation}
\end{footnotesize}
where $\Phi = [Z^{(v)}_{1}Z^{(v)}_{2}\ldots Z^{(v)}_{i-1}]$, and $\Phi$ is set as $I$ when $i=1$. Taking the derivative to zero and applying the Karush-Kuhn-Tucker (KKT) conditions, we can have:
\begin{equation}\label{eq:KKTZ}
\begin{aligned}
& \big(\Phi^TX^{(v)}(X^{(v)})^{\top}\Phi Z^{(v)}_{i} + \Phi^T\Phi Z^{(v)}_{i}H^{(v)}_{i}(H^{(v)}_{i})^{\top} -\\
& 2\Phi^{\top}X^{(v)}(H^{(v)}_{i})^{\top}\big)_{kl}\big( Z^{(v)}_{i}\big)_{kl}  - \big(\eta\big)_{kl}\big( Z^{(v)}_{i}\big)_{kl}  = 0.
\end{aligned}
\end{equation}
This process converges because this is a fixed point equation. Then we obtain the update rule as:
\begin{footnotesize}
\begin{equation}\label{eq:UZ}
\begin{aligned}
Z^{(v)}_{i} \! \leftarrow \! Z^{(v)}_{i}\!\odot \!
\frac{2\Phi^{\top}X^{(v)}(H^{(v)}_{i})^{\top}}{\Phi^{\top}\!X^{(v)}(X^{(v)})^{\top}\Phi Z^{(v)}_{i}\!\!+\!\Phi^{\top}\!\Phi Z^{(v)}_{i}H^{(v)}_{i}\!(H^{(v)}_{i}\!)^{\top}}.
\end{aligned}
\end{equation}
\end{footnotesize}
where $\odot$ represents the element-wise product.

\subsubsection{Update rule for $H^{(v)}_i (i<m)$:}
By utilizing a similar proof as~\cite{zhao2017multi}, we can formulate the update
rule for $H^{(v)}_i$ as follows:
\begin{footnotesize}
\begin{equation}\label{eq:UHi}
\begin{aligned}
\!(\!H^{\!(v\!)}_{i})\!_{kl}\!  \leftarrow \!(\!H^{(v)}_{i}\!)_{kl}\!\odot \!
\!\sqrt
{
\frac{\![2\Phi^{\top}X^{(v)}\!]_{kl}^{p} \!+\! [\Phi^{\top}\Phi H^{(v)}_i]_{kl}^{n} \!+\![H^{(v)}_i\!]_{kl}^{n}}
{\![2\Phi^{\top}X^{(v)}\!]_{kl}^{n} \!+\! [\Phi^{\top}\Phi H^{(v)}_i]_{kl}^{p} \!+\![H^{(v)}_i\!]_{kl}^{p}}
}.
\end{aligned}
\end{equation}
\end{footnotesize}

\subsubsection{Update rule for $H^{(v)}_m (i=m), G^{(v)}$ and $H^\ast$:}
Solving these variables is a challenging problem since it is hard to directly get the explicit solutions. We thus introduce two auxiliary variables
$M^{(v)}_1$ and $M^{(v)}_2$ to transform  the optimization Eq.~\eqref{eq:frameworkDetail}, and obtain the following objective function:
\begin{equation}\label{eq:UHm}
\begin{aligned}
\min \limits_{H_{m}^{(v)}} & \sum_{v=1}^V(\| X^{(v)}-\Phi H^{(v)}_{m} \|^2_{F}+ \| H^{(v)}_{m}-\Phi^{\top}X^{(v)} \|^2_{F} \\
&+ \beta \|M^{(v)}_1\|_{2,1} + \lambda \|M^{(v)}_2\|_{2,1})\\
s.t.,&~H^{(v)}_{m}\!\geq\! 0, M^{(v)}_1 \!=\! H^{(v)}_{m}A^{(v)}, M^{(v)}_2\!=\!H^{(v)}_{m} \!-\! G^{(v)}H^{\ast}.
\end{aligned}
\end{equation}
After converting Eq.~\eqref{eq:UHm} to an augmented Lagrangian function, we obtain the following expression:
\begin{equation}\label{eq:UHmLag}
\begin{aligned}
\mc{L}=& \sum_{v=1}^V(\| X^{(v)} - \Phi H^{(v)}_{m} \|^2_{F} + \| H^{(v)}_{m}- \Phi^{\top}X^{(v)} \|^2_{F} \\
&+\beta \|M^{(v)}_1\|_{2,1} + <\Lambda_1^{(v)},M^{(v)}_1-H^{(v)}_{m}A^{(v)}>  \\
& +\frac{\mu_1}{2} \| M^{(v)}_1 -  H^{(v)}_{m}A^{(v)} \|^{2}_{F} + \lambda \|M^{(v)}_2\|_{2,1}\\
&\; +<\Lambda_2^{(v)},M^{(v)}_2+G^{(v)}H^\ast-H^{(v)}_{m}> \\
&+ \frac{\mu_2}{2} \| M^{(v)}_2 -  H^{(v)}_{m}+ G^{(v)}H^\ast\|^{2}_{F} \\
& +<\Lambda_3^{(v)},-H^{(v)}_{m}+s> + \frac{\mu_3}{2} \|-H^{(v)}_m+s \|^{2}_{F} \\
s.t.,&~H^{(v)}_{m}\!\geq\! 0, M^{(v)}_1 \!=\! H^{(v)}_{m}A^{(v)}, M^{(v)}_2\!=\!H^{(v)}_{m} \!-\! G^{(v)}H^{\ast},
\end{aligned}
\end{equation}
where $\Lambda_1^{(v)}$, $\Lambda_2^{(v)}$ and $\Lambda_3^{(v)}$ are the Lagrangian multipliers,initialized with zero matrix; $\mu_1$, $\mu_2$ and $\mu_3$
are the parameters for penalty; $s>0$ is the slackness variable to satisfy the non-negative constraint for $H_m^{(v)}$. We then employ the alternating direction method of multipliers to solve this equation, and the update rules are as follows.

\textbf{Update rule for $H_m^{(v)}$:} With other variables fixed, we can
have the following Lagrangian objective function:
\begin{footnotesize}
\begin{equation}\label{eq:UHmLagF}
\begin{aligned}
\min \limits_{H_{m}^{(v)}} & \sum_{v=1}^V(\| X^{(v)} - \Phi H^{(v)}_{m} \|^2_{F} + \| H^{(v)}_{m}- \Phi^{\top}X^{(v)} \|^2_{F}  \\
&+ \!<\!\Lambda_1^{(v)},M^{(v)}_1-H^{(v)}_{m}A^{(v)}\!> \!+\! \frac{\mu_1}{2} \| M^{(v)}_1 \!-\!  H^{(v)}_{m}A^{(v)} \|^{2}_{F} \! \\
& +\!<\Lambda_2^{(v)},M^{(v)}_2+G^{(v)}H^\ast-H^{(v)}_{m}> + \frac{\mu_2}{2} \| M^{(v)}_2 -  H^{(v)}_{m} \\
& +G^{(v)}H^\ast\|^{2}_{F} +\! <\! \Lambda_3^{(v)},-H^{(v)}_{m}+s \!> \!+\! \frac{\mu_3}{2} \|\!-\!H^{(v)}_m+s \|^{2}_{F} \\
\end{aligned}
\end{equation}
\end{footnotesize}
Taking the derivative respect of $H^{(v)}_m$ to zero, we obtain:
\begin{equation}\label{eq:UHmS}
\begin{aligned}
&[2\Phi^{\top}\Phi + (\mu_2 +\mu_3 +2)I]H_m^{(v)} + \mu_1H_m^{(v)}A^{(v)}(A^{(v)})^\top = \\
&\;\;[\Lambda_1^{(v)}(A^{(v)})^\top + \Lambda_2^{(v)} +\Lambda_3^{(v)} + \mu_1\!M_1^{(v)}(A^{(v)})^\top  \\
&\;\;+ \mu_2 (M_2^{(v)}-G^{(v)}H^\ast) + \mu_3s].\\
\end{aligned}
\end{equation}
Since Eq.~\eqref{eq:UHmS} is a standard Sylvester equation, it can be effectively solved by Bartels-Stewart algorithm.

\textbf{Update rule for $G^{(v)}$ and $H^\ast$}: With other variables fixed except for $G^\ast$, we can have the following Lagrangian objective function:
\begin{equation}\label{eq:UGL}
\begin{aligned}
\min \limits_{G^{(v)}} & \sum_{v=1}^V <\Lambda_2^{(v)},M^{(v)}_2+G^{(v)}H^\ast-H^{(v)}_{m}> +\\
&\frac{\mu_2}{2} \| M^{(v)}_2 -  H^{(v)}_{m}+ G^{(v)}H^\ast\|^{2}_{F}.
\end{aligned}
\end{equation}
Taking the derivative to zero, we obtain the following update rule:
\begin{footnotesize}
\begin{equation}\label{eq:UGR}
\begin{aligned}
G^{(v)} \!=\! ((H^{(v)}_{m} \!-\! M^{(v)}_2)(H^\ast)^\top \!-\! \Lambda_2^{(v)}(H^\ast)^\top)(H^\ast(H^\ast)^\top))^\dag,
\end{aligned}
\end{equation}
\end{footnotesize}
where $\dag$ denotes the Moore-Penrose pseudo-inverse.

Similarly, $H^\ast$ can be updated with the following rule:
\begin{footnotesize}
\begin{equation}\label{eq:UH*R}
\begin{aligned}
H^{\ast} \!=\!  {((G^{(v)})\!^\top G^{(v)})}^\dag ((G^{(v)}\!)^\top(H^{(v)}_{m} \!-\! M^{(v)}_2)\!-\!  (G^{(v)})^\top\Lambda_2^{(v)}).
\end{aligned}
\end{equation}
\end{footnotesize}

\textbf{Update rule for $M_1^{(v)}$ and $M_2^{(v)}$: }
$M_1^{(v)}$ and $M_2^{(v)}$  are solved in a similar way as that to solve $G^{(v)}$, and we thus obtain the following update rules. The update rule for $M^{(v)}_1$ is written as follows:
\begin{equation}\label{eq:UM1R}
\begin{aligned}
M^{(v)}_1 = (\beta\Sigma_1+\mu_1I)^\dag(\mu_1H^{(v)}_{m}A^{(v)}-\Lambda_1^{(v)}),
\end{aligned}
\end{equation}
where $\Sigma_1$ is a diagonal matrix with the i-th diagonal element as ${(\Sigma_1)}_{ii} = \frac{1}{\|m^i\|^2}$. $m^i$ is the $i$-th row of the matrix $M_1^{(v)}$. $I$ is the identity matrix.

The update rule for $M^{(v)}_2$ can be written as follows:
\begin{equation}\label{eq:UM2R}
\begin{aligned}
M^{(v)}_2 = (\lambda\Sigma_2+\mu_2I)^\dag(\mu_2(H^{(v)}_{m}-G^{(v)}H^\ast-\Lambda_2^{(v)}).
\end{aligned}
\end{equation}

Until now, we have obtained all the update rules. We summarize the overall update process of the proposed
framework in \textbf{Algorithm~\ref{alg:updateALL}}.
\begin{algorithm}[t]
\caption{Optimization of Problem~\eqref{eq:frameworkDetail}}
\label{alg:updateALL}
\begin{algorithmic}[1]
\REQUIRE Visual-tactile data $\{X^{(v)}\}^{V}_{v=1}$,~layer size $p_i$,
\\hyper-parameter $\beta,\lambda,\mu_1,\mu_2,\mu_3$,~the number of clusters $k$\\
\STATE  \textbf{Initialize:}
\FOR{all layers in each modality}
\STATE $(Z_i^{(v)},H_i^{(v)})\leftarrow \mr{NMF}(H_{i-1}^{(v)})$
\STATE  $L^{(v)}\leftarrow$ $k$-NN graph construction on $X^{(v)}$
\ENDFOR
\WHILE {not converged}
\FOR{all layers in each modality}
\IF {$i<m$}
\STATE Update $H^{(v)}_i$ via Eq.~\eqref{eq:UHi}.
\ELSE
\STATE Update $H^{(v)}_m$(i.e., $H^{(v)}_i(i=m)$) via Eq.~\eqref{eq:UHmS}.
\STATE Update $G^{(v)}$ via Eq.~\eqref{eq:UGR}.
\STATE Update $H^\ast$ via Eq.~\eqref{eq:UH*R}.
\STATE Update $M_1^{(v)}$ via Eq.~\eqref{eq:UM1R}.
\STATE Update $M_2^{(v)}$ via Eq.~\eqref{eq:UM2R}.
\STATE Update Lagrangian multipliers $\Lambda_1^{(v)}$, $\Lambda_2^{(v)}$, $\Lambda_3^{(v)}$.
\ENDIF
\STATE Update $Z^{(v)}_i$ according to Eq.~\eqref{eq:UZ}.
\ENDFOR
\ENDWHILE
\RETURN $H^\ast$.
\end{algorithmic}
\end{algorithm}
After obtaining the optimized $H^\ast$, we could obtain the final clustering result by performing a standard spectral clustering on $H^\ast$.

\subsection{Time Complexity}
For the computational complexity, our proposed model consists of two steps, i.e., the pre-trained stage and the fine-tuned stage. In order to simplify the analysis, we suppose that all the layers are with the same size of hidden units. In the pre-trained stage, the computational complexity $\tau_p = \mc{O}(Vmt_p(dnp+pd^2+dn^2))$, where $V$ is the number of modalities, $m$ is number of layers, $p$ is the layer size, $d$ is the feature dimension, $n$ is the number of samples and $t_p$ is the number of iterations to achieve convergence in the pre-training process. In the fine-tuned stage, the  computational complexity is $\tau_f = \mc{O}(Vmt_f(dnp+pd^2+dn^2+p^3))$, where $t_f$ is the number of iterations. Thus, the total time complexity is $\tau_{total} = \tau_p+\tau_f$.

\section{Experiments}
In this section, we evaluate the performance of our proposed model via several empirical comparisons. We first provide the used datasets and experiment results, followed by some analyses about our model.

\subsection{Experimental Setting}
Extensive experiments are conducted on two visual-tactile fusion datasets and one benchmark dataset to evaluate our proposed model: \textbf{1) PHAC-2\footnote{http://people.eecs.berkeley.edu/~yg/icra2016}} dataset: it contains $8$ color images and $10$ tactile signals of $53$ household objects. In this paper, we utilize all images and the first 8 tactile signals. 4096-D visual and 2048-D tactile features are extracted in a similar way as~\cite{gao2016deep}.
\textbf{2) GelFoldFabric\footnote{http://people.csail.mit.edu/yuan\_{}wz/fabric-perception.htm}} dataset: it contains $10$ color images and $10$ tactile images of $118$ kinds of fabrics. More details about this dataset can be found in~\cite{yuan2017connecting}. In this paper, we use the pre-trained VGG-19 net to extract 4096-D features both for tactile and visual images.
\textbf{3)
Yale\footnote{http://vision.ucsd.edu/content/yale-face-database}} dataset: it is employed to evaluate the performance of the proposed framework when the modality number of the input data is more than 2, which contains $165$ images of $15$ subjects. Similar to~\cite{zhao2017multi}, three kinds of features (i.e., 3304-D LBP, 4096-D intensity, 6750-D Gabor) are extracted as different views.
\begin{table*}[!t]
\centering
\caption{Performance ($\%$) comparison of $6$ different metrics (mean $\pm$ standard deviation) on \textbf{PHAC-2} dataset.}
\begin{tabular}{|c|c|c|c|c|c|c|}
\hline
Method & ACC & NMI & AR & F-score & Precision & Recall \\
\hline\hline
Vision       & 35.14$\pm$1.89 &  64.73$\pm$1.35 &  10.81$\pm$2.68  & 13.18$\pm$2.49 & 8.35$\pm$0.24 & 13.24$\pm$1.48\\
\hline
Touch        & 26.25$\pm$1.03 &  55.97$\pm$0.79 &  7.52$\pm$0.80  & 9.34$\pm$0.72 & 7.91$\pm$0.85 & 11.48$\pm$0.63\\
\hline
ConcatFea    & 46.93$\pm$1.28 &  68.06$\pm$0.39 &  25.35$\pm$0.25  & 26.66$\pm$ 1.26 & 25.10$\pm$0.98 & 27.94$\pm$1.01\\
\hline
ConcatPCA    & 47.19$\pm$0.81 &  68.01$\pm$0.33 &  26.13$\pm$0.69  & 27.41$\pm$0.67 & 26.06$\pm$0.78 & 28.92$\pm$0.59\\
\hline
Co-Reg    & 50.98$\pm$0.20 &  61.05$\pm$0.51 &  15.31$\pm$0.63  & 16.81$\pm$0.62 & 15.75$\pm$0.58 & 18.04$\pm$0.66\\
\hline
Co-Training    & 52.30$\pm$1.70 &  72.36$\pm$1.30 &  32.37$\pm$1.90  & 32.30$\pm$3.00 & 33.52$\pm$2.90 & 36.74$\pm$2.90\\
\hline
Min-D    & 47.98$\pm$2.77 &  67.85$\pm$3.50 &  25.14$\pm$5.20  & 26.47$\pm$5.10 & 24.51$\pm$5.00 & 28.80$\pm$4.60\\
\hline
Multi-NMF    & 51.98$\pm$0.82 &  70.81$\pm$0.32 &  30.12$\pm$0.94  & 32.13$\pm$0.92 & 30.67$\pm$0.93 & 33.74$\pm$1.00\\
\hline
DiMSC    & 36.99$\pm$1.17 &  65.69$\pm$0.77 &  18.21$\pm$0.97  & 17.86$\pm$0.92 & 15.63$\pm$1.10 & 19.02$\pm$0.70\\
\hline
DMF-MVC   & 55.02$\pm$0.96 &  72.96$\pm$0.31 &  34.39$\pm$0.55  & 35.53$\pm$0.53 & 33.86$\pm$0.66 & 37.83$\pm$0.53\\
\hline
GLMSC   & 37.50$\pm$3.34 &  61.97$\pm$1.84 &  16.37$\pm$2.87  & 17.83$\pm$2.81 & 16.97$\pm$2.77 & 18.79$\pm$2.86\\
\hline
Ours& \textbf{59.17$\pm$1.40} &  \textbf{75.27$\pm$0.54} &  \textbf{38.97$\pm$1.13}  & \textbf{40.03$\pm$1.11} & \textbf{38.12$\pm$1.29} & \textbf{42.15$\pm$0.96}\\
\hline
\end{tabular}
\label{table2}
\end{table*}

\begin{table*}[htbp]
\centering
\caption{Performance ($\%$) comparison of $6$ different metrics (mean $\pm$ standard deviation) on \textbf{GelFabric} dataset.}
\begin{tabular}{|c|c|c|c|c|c|c|}
\hline
Method & ACC & NMI & AR & F-score & Precision & Recall \\
\hline\hline
Vision       & 35.46$\pm$1.08 &  65.91$\pm$0.70 &  17.30$\pm$1.26  & 17.96$\pm$1.25 & 16.87$\pm$1.17 & 19.21$\pm$1.34\\
\hline
Touch        & 33.92$\pm$1.05 &  65.00$\pm$0.52 &  15.71$\pm$0.92  & 16.39$\pm$0.91 & 15.42$\pm$0.85 & 17.48$\pm$1.00\\
\hline
ConcatFea    & 36.56$\pm$0.82 &  66.95$\pm$0.27 &  18.53$\pm$0.58  & 19.19$\pm$0.58 & 18.02$\pm$0.48 & 20.53$\pm$0.77\\
\hline
ConcatPCA    & 37.15$\pm$1.20 &  67.28$\pm$0.61 &  19.13$\pm$1.35  & 19.78$\pm$1.34 & 18.57$\pm$1.18 & 21.15$\pm$1.55\\
\hline
Co-Reg    & 45.80$\pm$1.28 &  55.33$\pm$0.47 &  36.09$\pm$0.68  & 36.54$\pm$0.70 & 33.39$\pm$0.88 & 39.63$\pm$0.78\\
\hline
Co-Training    & 37.85$\pm$0.78 &  45.85$\pm$0.78 &  35.14$\pm$1.70  & 35.59$\pm$1.74 & 32.43$\pm$2.00 & 39.27$\pm$1.62\\
\hline
Min-D    & 43.13$\pm$2.49 &  45.92$\pm$0.98 &  34.94$\pm$2.30  & 35.39$\pm$2.28 & 32.47$\pm$2.21 & 38.73$\pm$2.30\\
\hline
Multi-NMF    & 52.01$\pm$0.99 &  75.30$\pm$0.36 &  34.69$\pm$0.17  & 35.18$\pm$0.95 & 33.27$\pm$1.17 & 37.08$\pm$0.72\\
\hline
DiMSC    & 37.73$\pm$0.77 &  66.97$\pm$0.47 &  18.35$\pm$0.77  & 18.03$\pm$0.76 & 17.08$\pm$0.85 & 20.11$\pm$0.62\\
\hline
DMF-MVC   & 53.03$\pm$0.82 &  76.60$\pm$0.36 &  36.50$\pm$0.98  & 36.61$\pm$0.76 & 34.71$\pm$0.87 & 39.02$\pm$0.92\\
\hline
GLMSC   & 55.92$\pm$1.49 &  78.35$\pm$0.28 &  39.70$\pm$0.52  & 40.19$\pm$0.51 & 37.56$\pm$0.29 & 43.22$\pm$0.81\\
\hline
Ours& \textbf{62.19$\pm$0.55} &  \textbf{80.73$\pm$0.24} &  \textbf{45.86$\pm$0.65}  & \textbf{46.25$\pm$1.02} & \textbf{44.13$\pm$0.93} & \textbf{49.49$\pm$0.66}\\
\hline
\end{tabular}
\label{table3}
\end{table*}

\begin{table*}[!htbp]
\centering
\caption{Performance ($\%$) comparison of $6$ different metrics (mean $\pm$ standard deviation) on \textbf{Yale} dataset.}
\begin{tabular}{|c|c|c|c|c|c|c|}
\hline
Method & ACC & NMI & AR & F-score & Precision & Recall \\
\hline\hline
BestSV        & 61.60$\pm$3.00 &  65.40$\pm$0.90 &  44.00$\pm$1.10  & 47.50$\pm$1.10 & 45.70$\pm$1.10 & 49.50$\pm$1.00\\
\hline
ConcatFea    & 54.40$\pm$3.80 &  64.10$\pm$0.60 &  39.20$\pm$0.90  & 43.10$\pm$0.80 & 41.50$\pm$0.70 & 44.80$\pm$0.80\\
\hline
ConcatPCA    & 57.80$\pm$3.80 &  66.50$\pm$3.70 &  39.60$\pm$1.10  & 43.40$\pm$1.10 & 41.90$\pm$1.20 & 45.00$\pm$0.90\\
\hline
Co-Reg    & 56.40$\pm$0.20 &  64.80$\pm$0.20 &  43.60$\pm$0.20  & 46.60$\pm$0.00 & 45.50$\pm$0.40 & 49.10$\pm$0.30\\
\hline
Co-Training    & 63.00$\pm$0.10 &  67.20$\pm$0.60 &  45.20$\pm$1.00  & 48.70$\pm$0.09 & 47.00$\pm$1.00 & 50.50$\pm$1.62\\
\hline
Min-D    & 61.50$\pm$4.30 &  64.50$\pm$0.50 &  43.30$\pm$0.60  & 47.00$\pm$0.60 & 44.60$\pm$0.50 & 49.60$\pm$0.60\\
\hline
Multi-NMF    & 67.30$\pm$0.10 &  69.00$\pm$0.10 &  49.50$\pm$0.10  & 52.70$\pm$0.00 & 51.20$\pm$0.03 & 54.30$\pm$0.02\\
\hline
DiMSC    & 70.90$\pm$0.30 &  72.70$\pm$1.00 &  53.50$\pm$0.10  & 56.40$\pm$0.20 & 54.30$\pm$0.10 & 58.60$\pm$0.30\\
\hline
DMF-MVC   & 74.50$\pm$1.10 &  78.20$\pm$1.00 &  57.90$\pm$0.20  & 60.10$\pm$0.20 & 59.80$\pm$0.10 & 61.30$\pm$0.20\\
\hline
GLMSC   & 75.45$\pm$3.86 &  78.43$\pm$2.93 &  54.00$\pm$0.50   & 57.09$\pm$0.95 & 51.81$\pm$2.23 & 63.76$\pm$3.60\\
\hline
Ours& \textbf{80.73$\pm$0.63} &  \textbf{82.09$\pm$0.94} &  \textbf{64.51$\pm$0.69}  & \textbf{63.35$\pm$0.66} & \textbf{62.25$\pm$0.73} & \textbf{65.09$\pm$1.17}\\
\hline
\end{tabular}
\label{table4}
\end{table*}
\subsection{Comparison Models and Evaluation}
We compare our proposed framework with the following models including 7 multi-view baselines and 4 related single-view baselines.
\textbf{Related single-view clustering competitors:}
\textbf{Vision (Touch)} performs standard spectral clustering~\cite{ng2002spectral} on the visual (tactile) features;
\textbf{ConcatFea} concatenates all features first and then carries out standard spectral clustering; \textbf{ConcatPCA} concatenates all the features and does PCA to project the concatenated features into a low dimensional subspace, then performs standard spectral clustering on the projected features;
\textbf{Multi-view clustering competitors:}
\textbf{Co-Reg}~\cite{kumar2011coR} enforces the number shape between different views via co-regularizing the clustering hypotheses;
\textbf{Co-Training}~\cite{kumar2011coT} works on the hypothesis that the true underlying clustering would assign a point to the same cluster irrespective of the view;
\textbf{Min-D}~\cite{de2005spectral} creates a bipartite graph basing on the ``minimizing-disagreement" idea;
\textbf{Multi-NMF}~\cite{liu2013multi} utilizes non-negative matrix factorization to seek the common latent subspace for multi-view input data;
\textbf{DiMsc}~\cite{cao2015diversity} utilizes a diversity term to explore the complementary information of multi-view data;
\textbf{DNMF-MVC}~\cite{zhao2017multi} proposes a deep non-negative matrix factorization framework to capture the mutual information of multi-view data;
\textbf{GLMSC}~\cite{zhang2018generalized} simultaneously seeks the underlying representation and explores complementary information of multi-view data.

Similar to~\cite{cao2015diversity,zhao2017multi}, six different metrics i.e., accuracy~\textbf{(ACC)}, normalized mutual information~\textbf{(NMI)}, \textbf{Precision}, \textbf{F-score}, \textbf{Recall}, adjusted rand index ~\textbf{(AR)} are adopted to evaluate the clustering performance. Higher value indicates the better performance for all metrics. We run all algorithms $10$ times and report the mean values along with standard deviations.  Table~\ref{table2} and Table~\ref{table3} show the object clustering results on \textbf{PHAC-2} dataset and \textbf{GelFabric} dataset, respectively. Table~\ref{table4} shows the results on \textbf{Yale} dataset. \textbf{BestSV} performs standard spectral clustering on the features in each view and reports the best performance. For avoiding overfitting, the maximum number of iterations is set to 150 for all experiments.

From the presented results, we obtain the following observations: our framework achieves very competitive performance when comparing with all the competing models, which reveals the remarkable effectiveness of our framework in object clustering task. Specifically, the results shown in Table~\ref{table2} and Table~\ref{table3} reveal the importance of fusing visual and tactile information when comparing with the models using visual (or tactile) information alone. This observation also reveals that our framework is able to utilize the visual and tactile information more effectively, when comparing with state-of-the-arts. The results in Table~\ref{table4} also reveal that our framework is not limited to the $2$-modality (i.e., visual-tactile fusion) case, and it can be applied into other applications whose modality number is more than $2$.

\subsection{Ablation Study $\&$ Convergence Analysis}
In this subsection, we analyze the proposed framework from three perspectives. Firstly, we analyze the effectiveness of the proposed Auto-Encoder-like structure, graph regularization and the consensus regularization. Then, we analyze the parameter setting, followed by the convergence analysis.

\begin{figure}[!t]
\centering
\centerline{\includegraphics[width =.95\columnwidth]{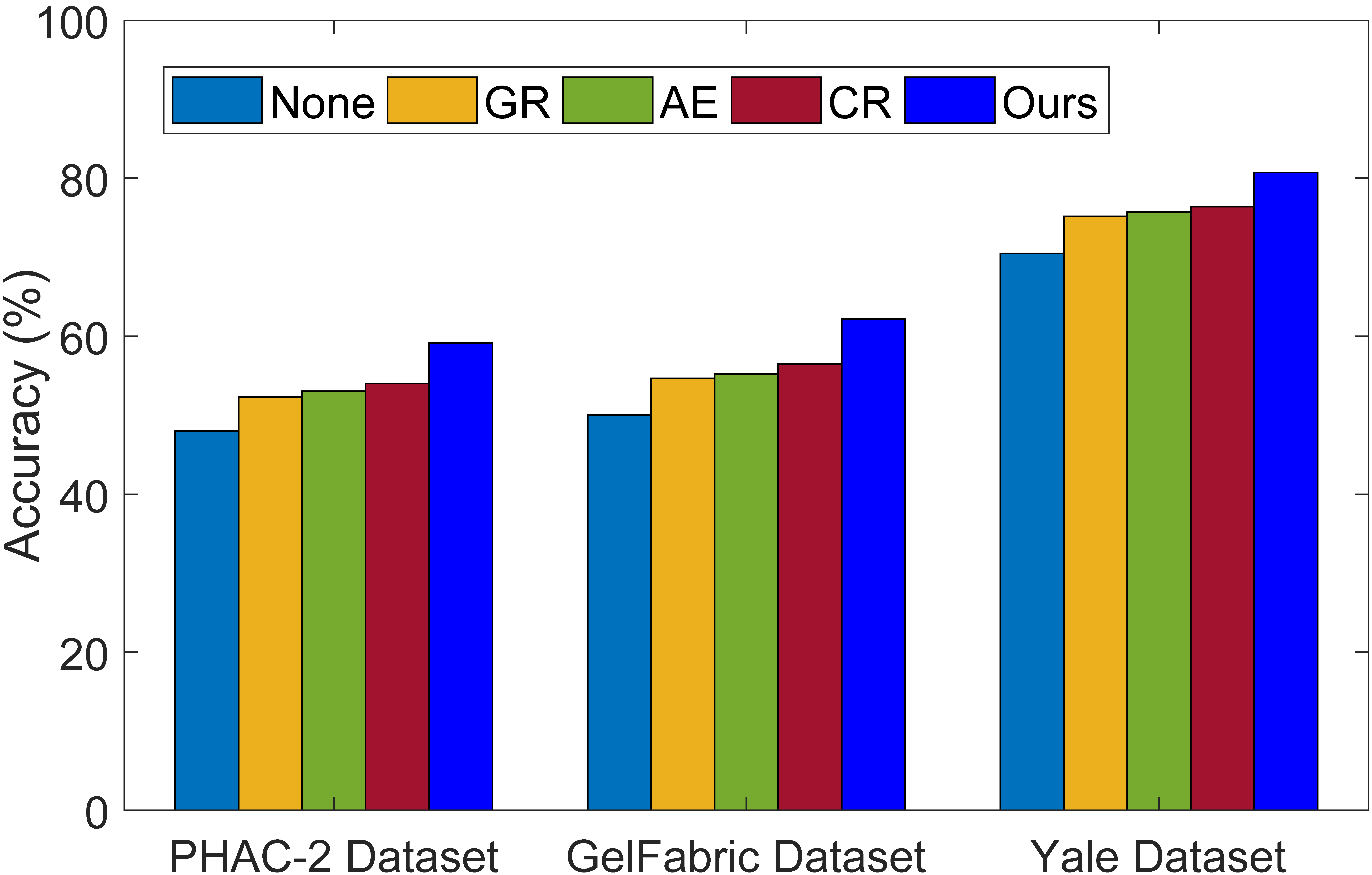}}
\caption{Effects of the Auto-Encoder-like structure, graph regularization and consensus regularization. ``None'' denotes that all items are not used while ``Ours" denotes that all items are used. ``AE'', ``GR'' and ``CR''  denote the models which only use the Auto-Encoder-like structure, the graph regularization, and the consensus regularization term, respectively.}
\label{fig:AECRGR}
\end{figure}

\textbf{Effectiveness of Auto-Encoder-like Structure, Graph Regularization and Consensus Regularization:} Figure~\ref{fig:AECRGR} presents the effectiveness of the used items. We can draw the following conclusion.
Overall, ``Ours'' achieves the best performance revealing that all the regularization and the Auto-Encoder-like structure proposed in this paper contribute to learn the rich information between multi-modality data which further boost the performance of  clustering tasks.
Specifically, ``AE'' achieve better performance than ``None'' denotes that via the proposed Auto-Encoder-like structure which takes data local structure preservation into account could result better representation for the source data.
``GR" achieve better performance than ``None" reveal the effectiveness of the graph regularization which can pull the similarities of nearby points and remove outliers inside each modality.
``CR'' achieve better performance than ``None" reveal that the proposed consensus regularization could  fill the gap between visual and tactile data and ultimately boost the clustering tasks.

\begin{figure}[!t]
\centering
\subcaptionbox{ACC(\%) curves w.r.t $\beta$}{\includegraphics[width =.49\columnwidth]
{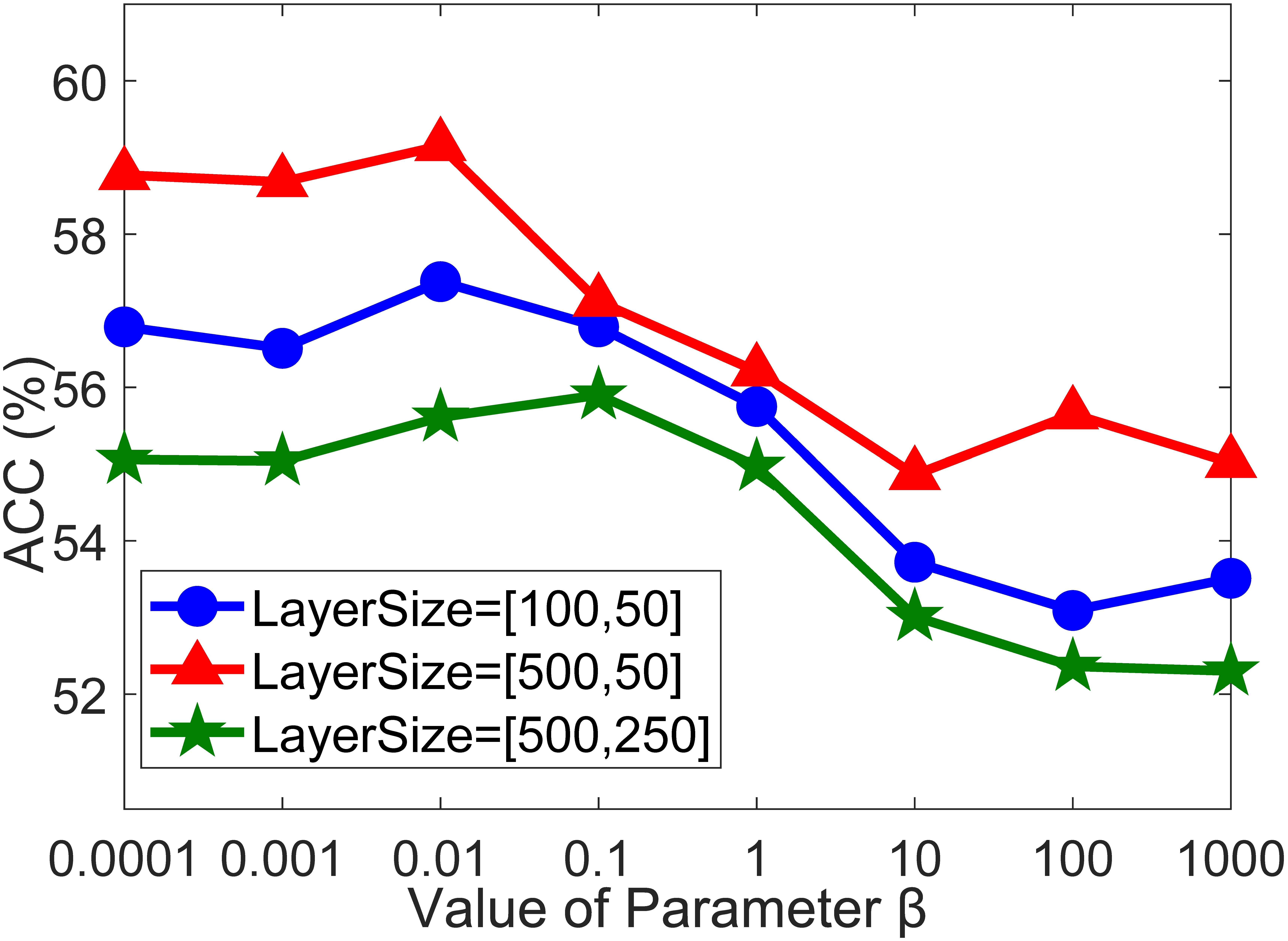}}
\subcaptionbox{NMI(\%) curves w.r.t $\beta$}{\includegraphics[width =.49\columnwidth]
{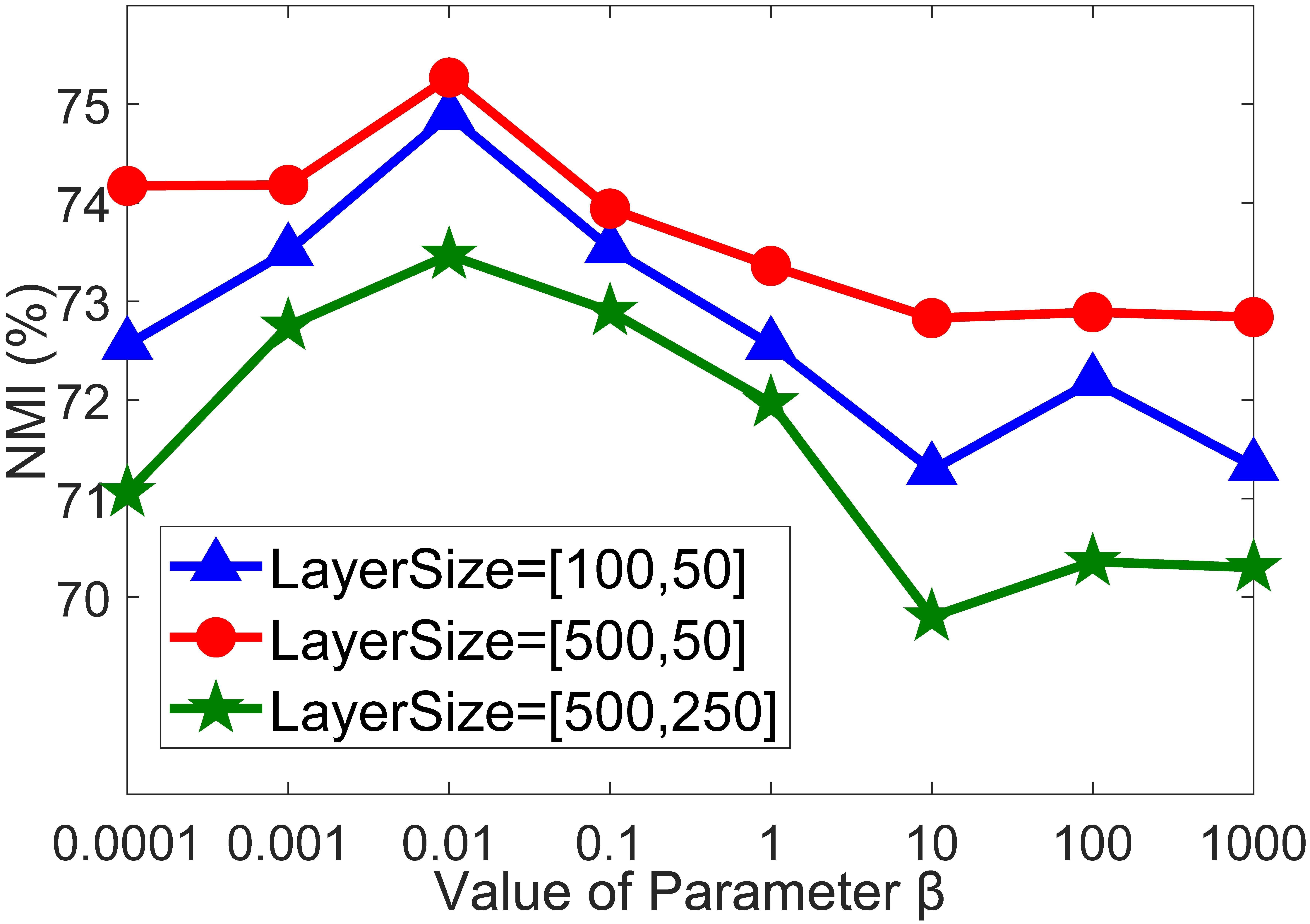}}
\caption{ACC(\%) and NMI (\%) curves w.r.t parameter $\beta$ on PHAC-2 dataset with different layer sizes. $\lambda$ is set as $0.1$.}
\label{fig:beta}
\end{figure}

\begin{figure}[t]
\centering
\subcaptionbox{ACC(\%) curves w.r.t $\lambda$}{\includegraphics[width =.49\columnwidth]
{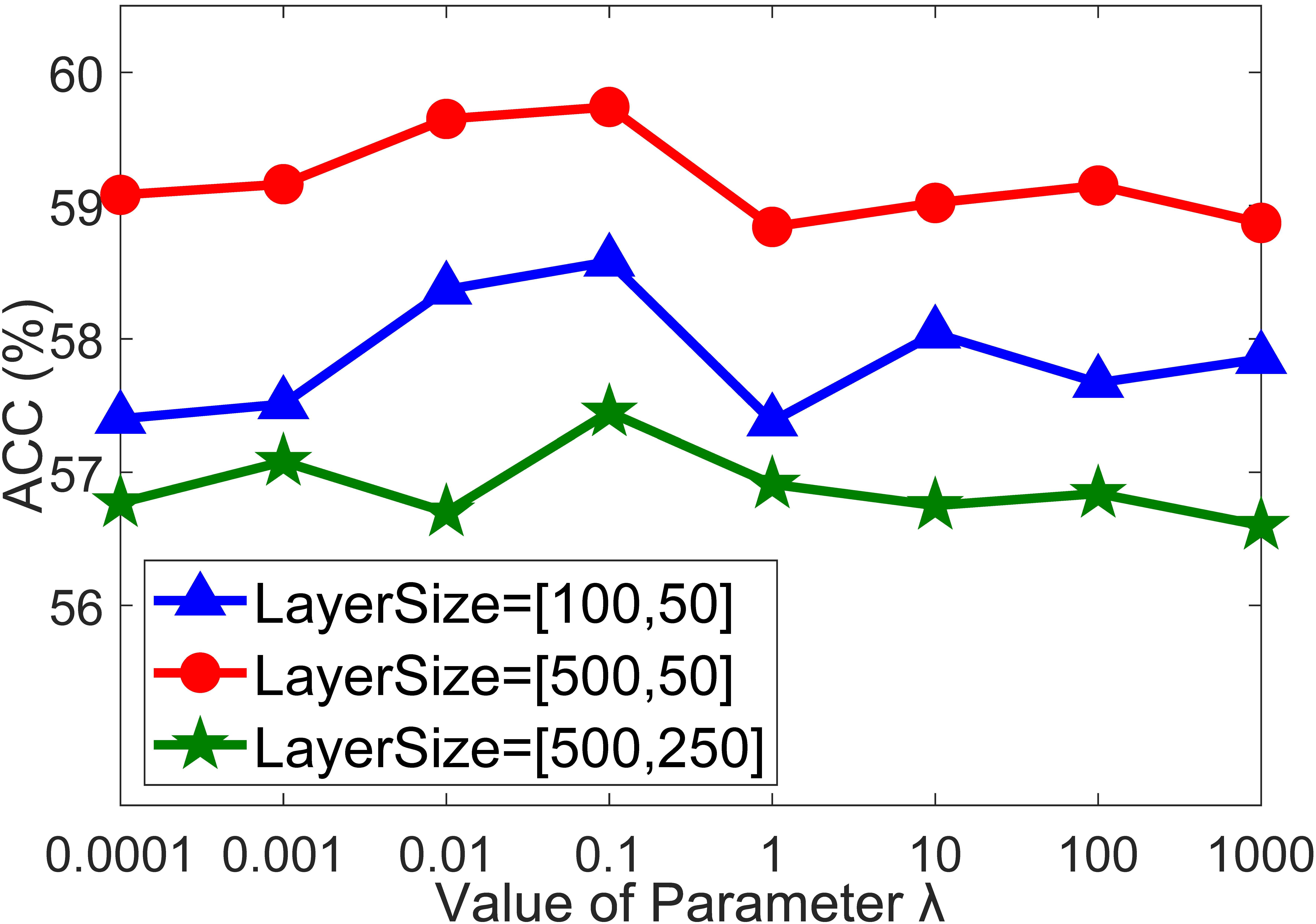}}
\subcaptionbox{NMI(\%) curves w.r.t $\lambda$}{\includegraphics[width =.49\columnwidth]
{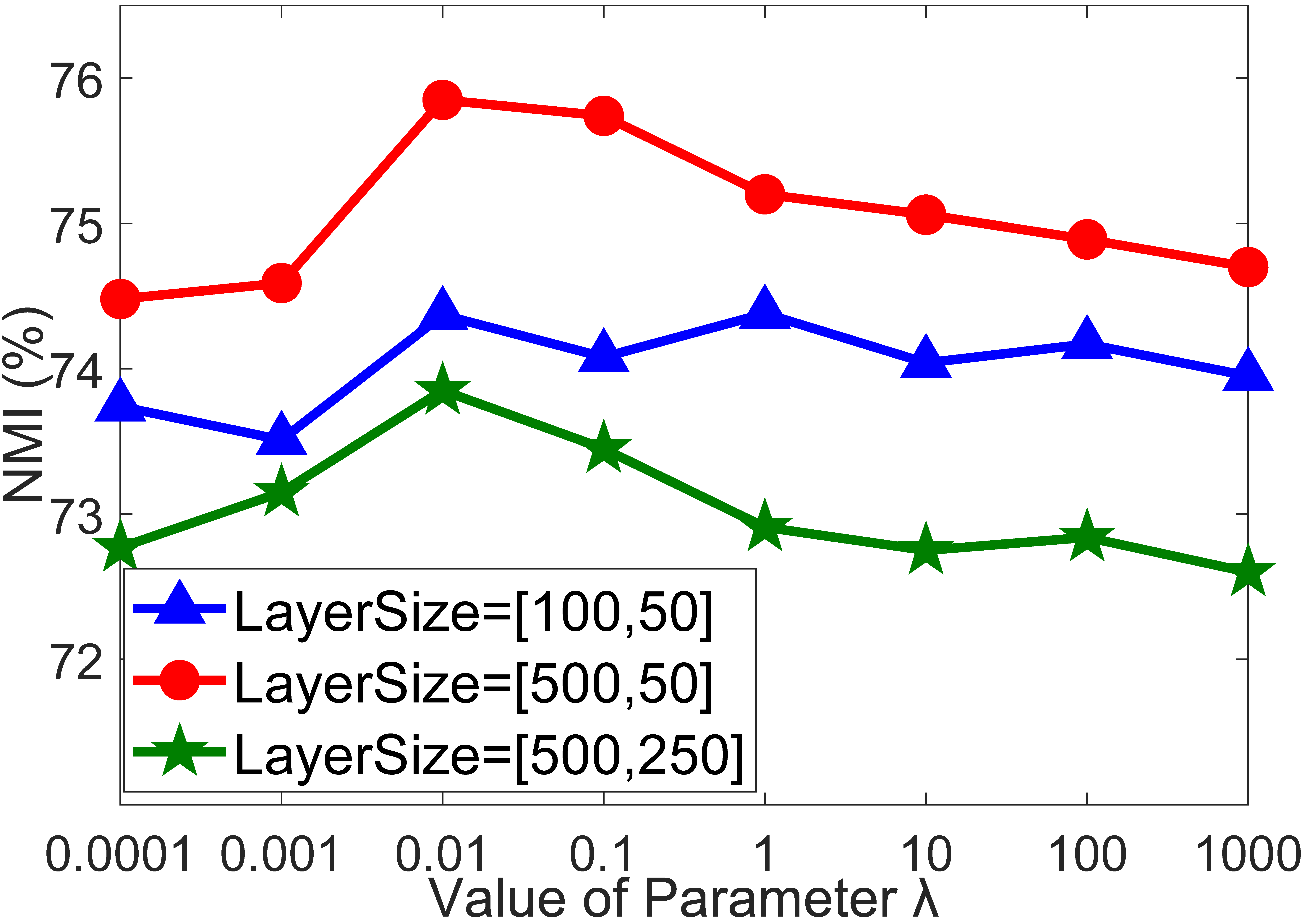}}
\caption{ACC(\%) and NMI (\%) curves w.r.t parameter $\lambda$ on PHAC-2 dataset with different layer sizes. $\beta$ is set as $0.1$.}
\label{fig:lambda}
\end{figure}

\textbf{Parameter Analysis:} To explore the effect of our used parameters, i.e, control parameters $\lambda$ and $\beta$ and the layer size $p_i$, we use \textbf{PHAC-2} dataset in this subsection. Specifically, Figure~\ref{fig:beta} shows the influence of ACC and NMI results w.r.t. the parameter $\beta$ under different layer sizes. As can be seen, under three different layer sizes, the framework performs best both in ACC and NMI when $\beta$ is set as $0.01$. We thus set $\beta=0.01$ as default in this paper. Figure~\ref{fig:lambda} explores the parameter sensitivity of the proposed framework w.r.t. the parameter $\lambda$ under different layer sizes. In this experiment, $\beta$ is set as $0.01$. Notice that the framework perform best both in ACC and NMI when $\lambda$ is set as $0.01$. So $\lambda=0.01$ is set as default. Figure~\ref{fig:beta} and Figure~\ref{fig:lambda} also explore the influence of model performance w.r.t. the layer sizes. We find that the setting of $[500,50]$ always leads to best performance. When the layer size is small, the framework is insufficient to learn the rich information behind the input data. And when the layer size is too large, it might introduce undesirable noise. This might be the possible reason why red curves perform better (i.e, layer size is $[500,50]$ ) than the blue curves (i.e.,$[100,50]$)and the green curves (i.e.,$[500,250]$).

\begin{figure}[htbp]
\centering
\subcaptionbox{PHAC-2 Dataset}{\includegraphics[width =.49\columnwidth]
{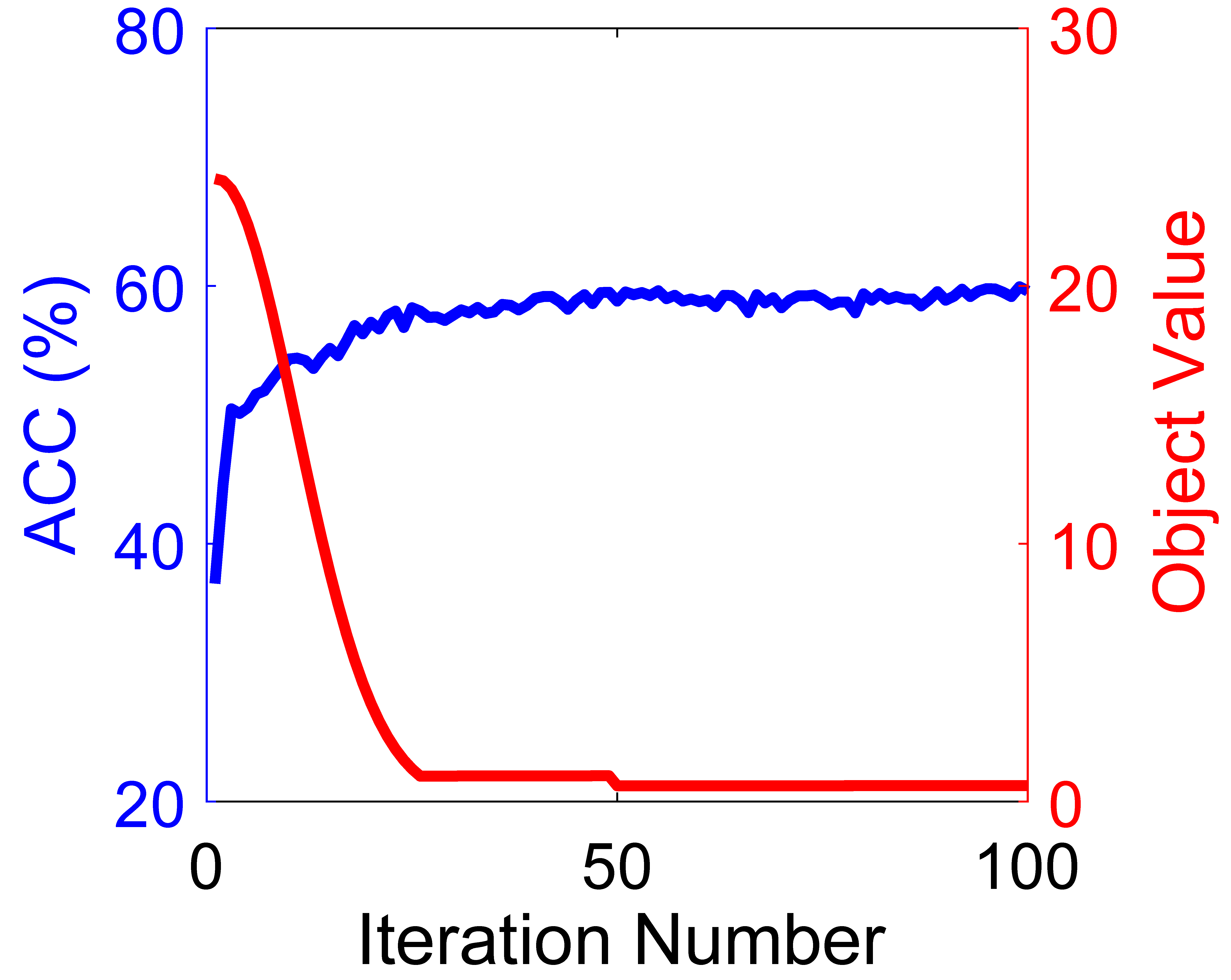}}
\subcaptionbox{GelFabric Dataset}{\includegraphics[width =.49\columnwidth]
{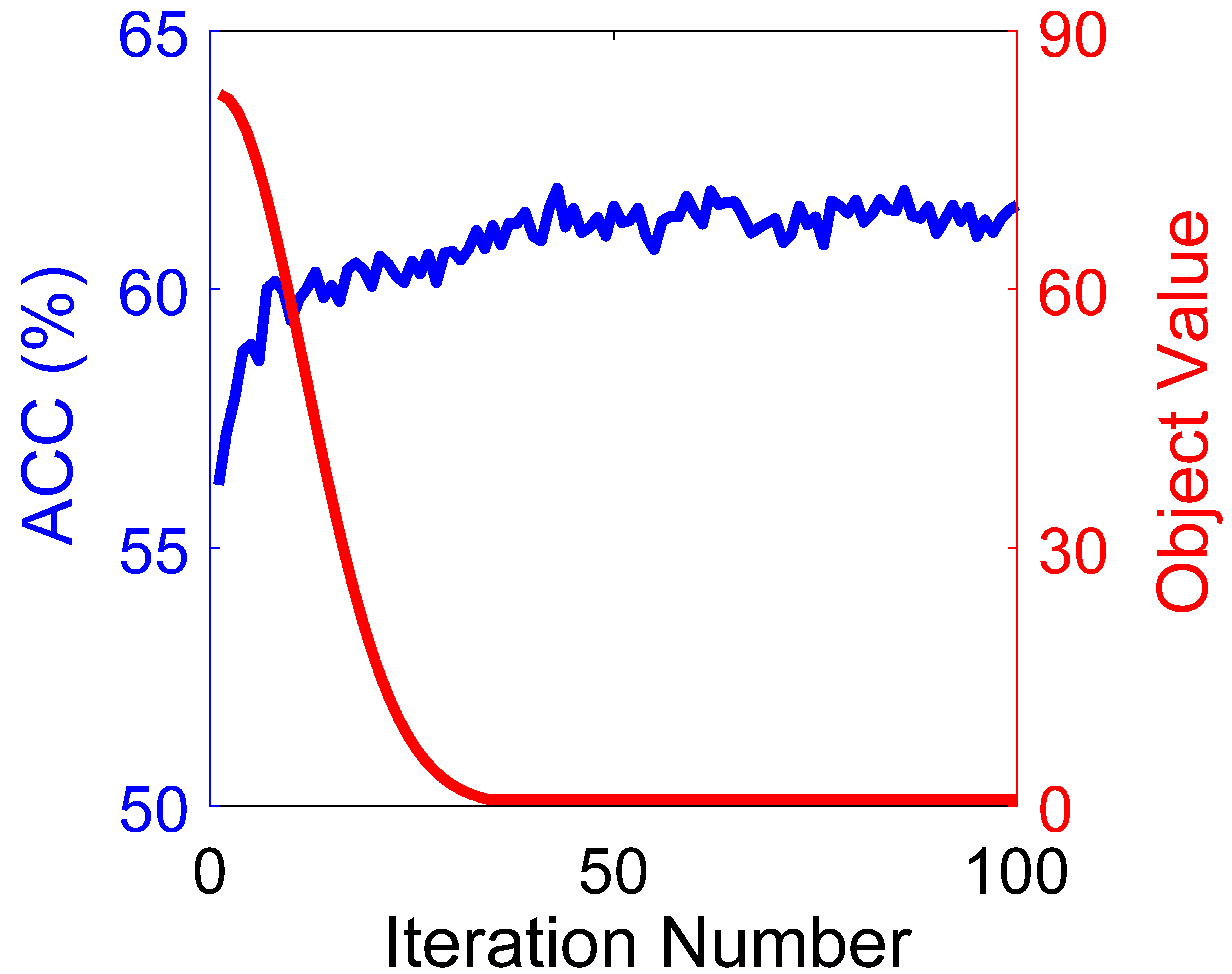}}
\caption{Convergence analysis on PHAC-2 dataset (a) and GelFabric dataset (b). ACC(\%) (blue line) and objective function value (red line) w.r.t. iteration time, respectively.}
\label{fig:conver}
\end{figure}
\textbf{Convergence Analysis:} Even though we have not proved that the proposed framework theoretically converges, we present the convergence property empirically in Figure~\ref{fig:conver}.
The objective value and ACC are plotted and we choose the default parameters, i.e., $\beta = 0.01$, $\lambda = 0.1$ and layer size = $[500,50]$ in this experiments.
Notice that the objective value gradually decreases until it converges after $100$ iterations.
ACC has two stages: in the first stage, ACC increases rapidly; in the second stage, ACC grows slowly and sightly bumps until reaching the best performance.

\section{Conclusion}
In this paper, we propose a deep Auto-Encoder-like NMF framework for visual-tactile fusion object clustering.
By constraining the deep NMF architecture by an under-complete Auto-Encoder-like structure, our framework can jointly learn the hierarchical semantics of visual-tactile data and maintain the local structure of the source data. For each modality, a graph regularization is adopted to pull the similarities of nearby points and remove outliers inside each modality. To create a common subspace in which the gap between visual and tactile data is filled, a sparse consensus regularization is developed in this paper, while the mutual information amongst visual and tactile data is maximized. Extensive experiment results on two visual-tactile fusion datasets and one benchmark dataset confirm the effectiveness of our framework, comparing with existing state-of-the-art works.

\bibliographystyle{aaai}
\bibliography{596}
\end{document}